\documentclass[conference]{IEEEtran}
\IEEEoverridecommandlockouts
\usepackage{cite}
\usepackage{amsmath,amssymb,amsfonts}
\usepackage{algorithmic}
\usepackage{graphicx}
\usepackage{textcomp}
\usepackage{xcolor}

\usepackage{booktabs} % For \toprule, \midrule, \bottomrule
\usepackage{multirow}
\usepackage{tabularx}
\usepackage{array}

\definecolor{bluehighlight}{rgb}{0,0,1}
\definecolor{redcross}{rgb}{1,0,0}

\usepackage[pagebackref=false,breaklinks,colorlinks,bookmarks=false]{hyperref}
\usepackage[capitalize]{cleveref}
\crefname{section}{Sec.}{Secs.}
\Crefname{section}{Section}{Sections}
\Crefname{table}{Table}{Tables}
\crefname{table}{Tab.}{Tabs.}

\def\BibTeX{{\rm B\kern-.05em{\sc i\kern-.025em b}\kern-.08em
    T\kern-.1667em\lower.7ex\hbox{E}\kern-.125emX}}

\begin{document}
\bstctlcite{IEEEexample:BSTcontrol} 

\title{Attention-guided Evidence Grounding for Spoken Question Answering}

% Double-blind review: Do not include author names
\author{
\IEEEauthorblockN{
Ke Yang, Bolin Chen, Yuejie Li, Yueying Hua, \\
Jianhao Nie, Yueping He, Bowen Li, Chengjun Mao
}
\IEEEauthorblockA{
\textit{Ant Group}\\
Hangzhou, China \\
\{zhulang.yk, bolin.cbl, liyuejie.lyj, huayueying.hyy, niejianhao.njh, heyueping.hyp, zhikong.lbw, chengjun.mcj\}@antgroup.com
}
}

\maketitle

\begin{abstract}
Spoken Question Answering (Spoken QA) presents a challenging cross-modal problem: effectively aligning acoustic queries with textual knowledge while avoiding the latency and error propagation inherent in cascaded ASR-based systems. In this paper, we introduce Attention-guided Evidence Grounding (AEG), a novel end-to-end framework that leverages the internal cross-modal attention of Speech Large Language Models (SpeechLLMs) to explicitly locate and ground key evidence in the model’s  latent space. To address the diffuse attention distribution in pre-trained models, we propose Learning to Focus on Evidence (LFE), a supervised fine-tuning paradigm that calibrates the model's attention mechanism to distinguish query-relevant segments from irrelevant context. 
Experiments on SQuAD, HotpotQA, and MuSiQue demonstrate that AEG reduces hallucinations and achieves strong efficiency gains, outperforming large-scale cascaded baselines (Whisper-Large-v3 + Reranker) while reducing inference latency by approximately 62\%.
% Extensive experiments on SQuAD, HotpotQA, and MuSiQue benchmarks demonstrate that AEG not only mitigates factual inconsistencies (hallucinations) but also establishes a new state-of-the-art in efficiency. Specifically, AEG achieves an F1 score of 80.02\% on evidence grounding, outperforming large-scale cascaded baselines (Whisper-Large-v3 + Reranker) while significantly reducing inference latency by approximately 62\%.
\end{abstract}

\begin{IEEEkeywords}
Spoken Question Answering, Speech Large Language Models, Cross-modal Attention, End-to-End Systems
\end{IEEEkeywords}

\section{Introduction}
\label{sec:intro}

Spoken Question Answering (Spoken QA) \cite{chi-etal-2025-role,shon2024discreteslulargelanguagemodel} is a cross-modal task that requires models to answer spoken queries based on textual contexts. However, even when provided with the correct context, current Spoken QA systems frequently generate responses that are inconsistent with the source content, resulting in hallucinations \cite{Huang_2025, zhang2025siren}. This severely limits their deployment in critical, high-stakes scenarios such as medicine \cite{pal2023med} and legal consultation \cite{dahl2024large}.
% such as medicine \cite{pal2023med,granstedt2025hallucinations}
% and legal consultation \cite{dahl2024large, magesh2025hallucination}.
% Furthermore, existing approaches lack explicit mechanisms for evidence identification and grounding, which reduces the interpretability of current models. Consequently, user cannot verify which parts of the context support the generated answer.
Furthermore, existing approaches lack explicit evidence grounding, limiting interpretability and preventing user verification of supporting contexts.
Therefore, improving the factual accuracy and interpretability of spoken QA systems is crucial for their reliable deployment in real-world applications.

Inspired by human cognitive processes, our work explores a more intrinsic and intuitive approach to spoken QA. When humans perform a similar task, they intuitively follow a "scan-then-focus" process: Given a large amount of context, they first scan and identify key information segments relevant to the question—i.e., the key evidence. Subsequently, they focus their attention on these key evidence to construct an answer, rather than attempting to process all context simultaneously. This two-stage process helps to reduce irrelevant interference during reasoning and ensures that the answer is based on verified information.

We posit that a similar mechanism can be instilled in Speech Large Language Models (SpeechLLMs). Intuitively, as an LLM processes a query and its context, its internal attention mechanism is already dynamically computing the importance of different information segments \cite{vaswani2017attention, hassid2022much}, creating a theoretical "heat map" of relevance. Hence, attention offers a promising but largely under-exploited signal for evidence grounding \cite{zhang2024selective, leviathan2024selective}. Motivated by this insight, we first propose \textbf{Grounding with Attention}, which leverages the internal attention mechanism of a pre-trained SpeechLLM to locate and explicitly mark key evidence. These marked segments provide clear attribution for the model’s responses, enhancing interpretability by revealing the specific context used during answer generation.

\begin{figure*}[t]
\centering
\includegraphics[width=0.95\textwidth]{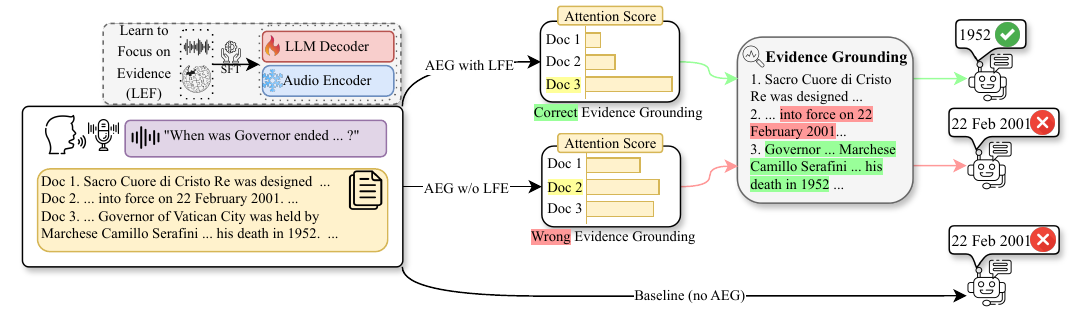}
\caption{A comparison demonstrating the critical role of Learning to Focus on Evidence (LFE). For the audio query "When was the Governor ended?", our complete AEG framework successfully grounds the answer in the correct evidence (Doc 3), while both AEG without LFE and baseline methods fail to identify relevant evidence, resulting in incorrect responses.} 
\label{fig:intro-case}
\end{figure*}

While attention-based grounding works well in text-only settings, it is unreliable in cross-modal speech–text scenarios, where attention must align heterogeneous acoustic and text representations and thus requires explicit training.
As a result, the raw attention of pre-trained SpeechLLMs is often diffuse and uncalibrated, failing to clearly distinguish key evidence from related context, as illustrated in \Cref{fig:intro-case}.
% However, we also observe that the raw, untrained attention of a pre-trained SpeechLLM is often diffuse and uncalibrated. The attention distribution tends to be relatively uniform across all context passages, failing to sharply distinguish key evidence from merely related information. This limitation is clearly illustrated by our case study in \Cref{fig:intro-case}. 
To address this issue, we propose \textbf{Learning to Focus on Evidence (LFE)}, a specialized training paradigm designed to "teach" the LLM to perform the "scan-then-focus" process. Through task-specific fine-tuning, LFE reshapes the model's attention distribution, producing a sharper distinction between key evidence and irrelevant context. This enables the model to concentrate more effectively on key evidence. 

Overall, we present \textbf{Attention-guided Evidence Grounding (AEG)}, a controllable framework that starts from grounding with attention and provides a trainable way to guide a model's focus on relevant evidence. By transforming the model's implicit attention patterns into explicit evidence markers, AEG significantly enhances both the factual accuracy and the interpretability of the generated responses. 

% Our main contributions are as follows: 
% \begin{itemize}
% \item We propose \textbf{Attention-guided Evidence Grounding (AEG)}, a framework that leverages the internal attention mechanism of a SpeechLLM to identify key evidence segments in the given context, and guide the answer generation process.
% \item We introduce \textbf{Learning to Focus on Evidence (LFE)}, a specialized training paradigm that fine-tunes the model on a selective generation task to sharpen its attention on key evidence.
% \item Experiments on multiple Spoken QA benchmarks demonstrate that \textbf{AEG} significantly improves evidence selection precision, enhances factual correctness in generated answers. 
% \end{itemize}

Our main contributions are as follows:
\begin{itemize}
    \item We propose \textbf{AEG}, a framework that leverages the internal attention mechanism of SpeechLLMs to explicitly locate key evidence within the context, thereby guiding the model to generate well-grounded answers.
    \item We introduce \textbf{LFE}, a specialized fine-tuning paradigm that calibrates the model's attention distribution. This process sharpens the focus on critical information, enabling the model to filter out irrelevant noise effectively.
    \item Experiments on multiple Spoken QA benchmarks demonstrate that AEG significantly improves evidence selection precision and enhances factual correctness, effectively boosting the overall reliability of the system against hallucinations.
\end{itemize}

\section{Related Work}

Building on the success of Large Language Models (LLMs) in the text domain, a natural next step is extending this paradigm to other modalities, particularly speech. Traditionally, tasks like Spoken Question Answering have used cascaded ASR-LLM-TTS architectures \cite{10832300}, but this approach suffers from error propagation, high latency, and loss of paralinguistic information. 
To address these limitations, the research trend is shifting towards integrated, end-to-end (E2E) SpeechLLMs \cite{He2025SurveyOE,SETHIYA2025101751,team2025longcat,tian2025step,Inclusion2025MingOmniAU,li2026squtrrobustnessbenchmarkspoken}. These models directly process speech input and generate speech or text output, enabling a deeper fusion of acoustic and semantic information.

Despite advancements in E2E architectures, SpeechLLMs still face the persistent challenge of "hallucination": the generation of outputs are plausible but factually incorrect or unfaithful to the context. Although Retrieval-Augmented Generation (RAG) can mitigate this issue, the "faithfulness" of the model to the given context remains a critical concern. Research indicates that even when the correct documents are retrieved, models may still ignore context \cite{gao-etal-2023-enabling}, become "confused" by contradictory information, or "selectively" use information. Furthermore, model efficacy is often hampered by the "lost in the middle" phenomenon \cite{Liu2023LostIT}, where LLMs exhibit a strong positional bias, favoring information at the beginning and end of the context while overlooking critical evidence in the middle.

Inspired by these challenges, recent research has explored more fine-grained contextual utilization, predominantly within the text modality. Relevant efforts include re-ranking documents based on attention scores to counteract positional bias \cite{Peysakhovich2023AttentionSC}, employing auxiliary evidence extractors \cite{Wang2023LearningTF}, or leveraging self-attention mechanisms to highlight evidence \cite{Liu2025SelfElicitYL}. However, these methods are limited to processing purely textual information. They fail to address a critical issue in multimodal models: how to effectively align and ground the user's acoustic information with external textual knowledge. In contrast, our work focuses on achieving direct alignment between speech queries and textual knowledge passages within multimodal models.

\section{Methodology}

\subsection{Task Definition}
We formally define the spoken question answering task as follows.
Let $Q_A$ be an audio query and $C_T=\{c_1,c_2,...,c_k\}$ be a context set, which serves as the exclusive knowledge source for answering the query. The task takes the audio query $Q_A$ and context set $C_T$ as inputs, which are processed by a speech large language model (SpeechLLM) to generate a textual response $R_T$:
\begin{equation}
    R_T=SpeechLLM(Q_A, C_T).
\end{equation}

\subsection{Overall Framework}

\begin{figure}[t]
\centering
\includegraphics[width=0.48\textwidth]{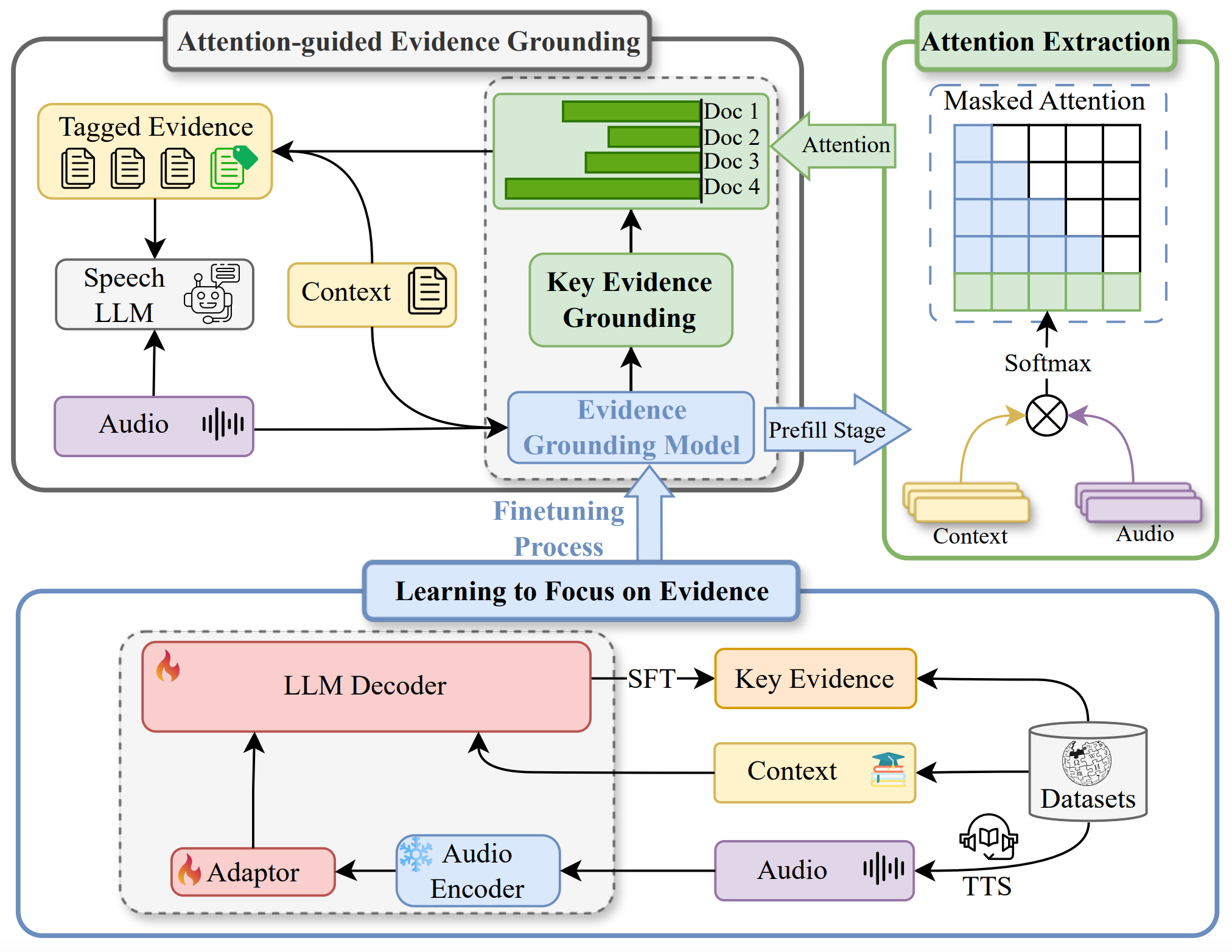}

\caption{Overview of the proposed \textbf{Attention-guided Evidence Grounding (AEG)} method. AEG comprises two components: (1) \textbf{Learning to Focus on Evidence}—a supervised fine-tuning stage that calibrates the SpeechLLM’s attention toward key evidence, and (2) \textbf{Grounding with Attention}—an inference stage that leverages learned attention patterns to highlight and ground key evidence.}
\label{fig:architecture}
\end{figure}

The proposed Attention-guided Evidence Grounding (AEG) method, illustrated in \Cref{fig:architecture}, presents a novel framework for spoken question answering that leverages the internal attention mechanism of a SpeechLLM to effectively identify and ground key evidence.

The inference pipeline begins with a spoken query $Q_A$ and a set of textual context segments $C_T=\{c_1,c_2,...,c_k\}$. 
These inputs are processed by the \textbf{Evidence Grounding Model}, and attention weights are computed, typically during the prefill stage. The \textbf{Attention Extraction} module then derives the attention map (visualized as Masked Attention), reflecting the importance of each segment $c_i$.
Based on these attention weights, the model identifies and selects a subset of the most important contexts, which we refer to as \textbf{key evidence}. To explicitly guide the model’s response generation, we "ground" this evidence by annotating it with special markers (e.g., $\texttt{<EVIDENCE>}$ and $\texttt{</EVIDENCE>}$). Finally, the original query $Q_A$ and the annotated context $C_{T}^{'}$ are passed to the SpeechLLM to generate the final response $R_T$.

However, we observe that current base models exhibit relatively uniform attention distributions across context segments. To improve this, we introduce a specialized training method, \textbf{Learning to Focus on Evidence (LFE)}. This method fine-tunes the SpeechLLM to enhance its attention mechanism, enabling it to better distinguish query-relevant contexts from irrelevant ones. This training process refines the initial SpeechLLM into an optimized \textbf{Evidence Grounding Model}.

\subsection{Grounding with Attention}
Large language models (LLMs) operate in two distinct phases during inference: prefill and decode. In our approach, we extract and analyze attention weights from the prefill phase to locate and mark key evidence that is most relevant to answering the query.

\subsubsection{Attention Weight Extraction}
Given a spoken query $Q_A$ and the textual context set $C_T=\{c_1,c_2,...,c_k\}$, we first process them into a unified input sequence $S$. The audio query $Q_A$ is transformed via the $\text{AudioEncoder}$, and the textual context $C_T$ is processed using the $\text{Tokenizer}$. The complete input sequence $S$ is then constructed by concatenating these representations:
\begin{equation}
\label{eq:input_seq}
S = AudioEncoder(Q_A) \oplus Tokenizer(C_T).
\end{equation}

Our goal is to calculate the importance score for each context segment through a hierarchical aggregation of self-attention weights. Denote by ${\alpha}^{(l, h)}_{j}$ the attention weight at layer $l \in {1,...,L_{end}}$ and head $h \in {1,...,H}$. For each token $t_j$ belonging to context $c_i$, the head-averaged attention in layer $l$ is:
\begin{equation}
\label{eq:token_attention_head}    
A^{(l)}(t_j) = \frac{1}{H} \sum_{h=1}^{H} \boldsymbol{\alpha}^{(l, h)}_{j}.
\end{equation}

Next, we aggregate across a selected range of layers. Based on empirical analysis (detailed in Section \ref{sec:attention_layer_analysis}), we identify a contiguous block of layers, from $L_{start}$ to $L_{end}$, where attention patterns are most indicative of semantic relevance. The attention weights for token $t_j$ is the average of its weights across these layers:
\begin{equation}
\label{eq:token_attention_layer}    
\bar{A}(t_j) = \frac{1}{L_{end}-L_{start}+1} \sum_{l=L_{start}}^{L_{end}} A^{(l)}(t_j).
\end{equation}

To obtain attention weights of the context segment, we aggregate token-level attention weights within each context segment. The overall importance score for a context segment $c_i$, denoted $A(c_i)$, is calculated as the mean of the final weights of all its constituent tokens:
\begin{equation}
\label{eq:passage_attention}
A(c_i) = \frac{1}{|c_i|} \sum_{t_j \in c_i} \bar{A}(t_j).
\end{equation}

We apply a threshold-based selection to determine the set of key evidence:
\begin{equation}
\label{eq:thresholding}
C_{\text{key}} = {c_i \in C_T \mid A(c_i) > \tau},
\end{equation}
where $\tau$ is a predefined threshold, and $C_{\text{key}}$ contains the set of candidates whose attention weights $A(c_i)$ exceed $\tau$.

\subsubsection{Key Evidence Grounding} For each context segment $c_i \in C_{key}$, reserved marking tokens (e.g., $\texttt{<EVIDENCE>}$ and $\texttt{</EVIDENCE>}$) are inserted before and after its content, producing a tagged context set $C_T'$. The original query $Q_A$ and the tagged context $C_T'$ are then provided to a generative QA LLM:
\begin{equation}
R_T = SpeechLLM(Q_A, C_T').
\end{equation}

\begin{figure*}[t]
\centering
\includegraphics[width=1.0\textwidth]{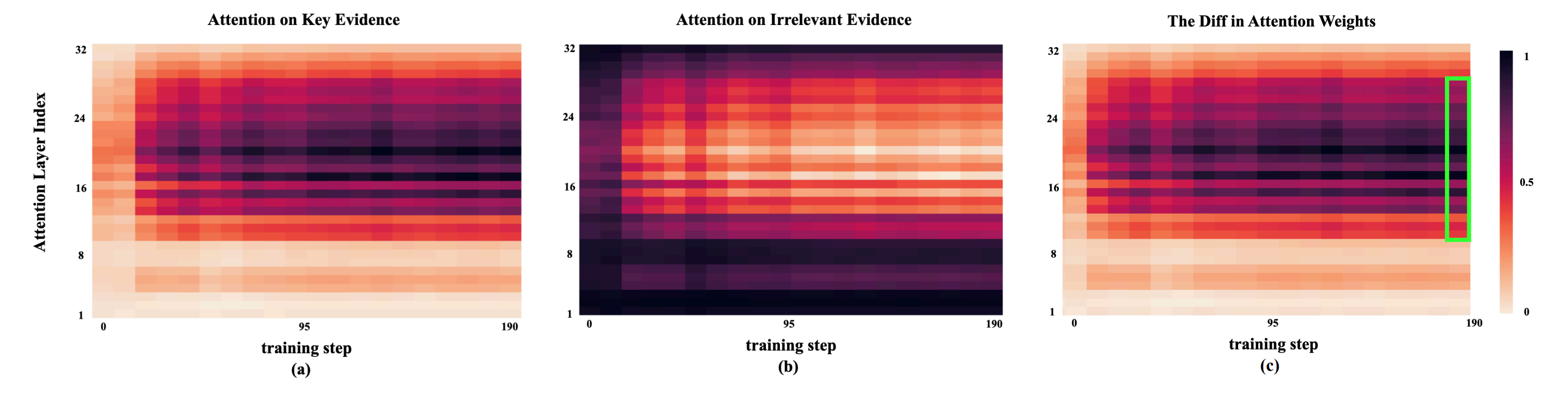}
\caption{Heatmaps of attention weight evolution across layers (y-axis, 1-32) and training steps (x-axis, 0-190) in Qwen2-Audio. (a) Weights allocated to key evidence. (b) Weights assigned to irrelevant evidence. (c) The difference (diff) between key and irrelevant weights. Green boxes (10-28) highlight the most effective layers.} \label{fig:attention_evolution}
\end{figure*}

\subsection{Learning to Focus on Evidence }
\label{sec:LFE}
As shown in Eq. \ref{eq:passage_attention}, the passage-level attention weight $A(c_i)$ is computed by averaging token-level weights. However, unmodified attention mechanisms often dilute high weights in long segments or overweight short irrelevant segments.

To address this limitation, we fine-tune the SpeechLLM model via supervised fine-tuning (SFT) to enhance its ability to discriminate key evidence. We formulate the SFT stage as a selection generation task, training the model to generate only the ground-truth evidence. By forcing the model to reconstruct only the key evidence, the auto-regressive loss function naturally penalizes attention to non-evidence tokens during generation.

Given a spoken query $Q_A$ and a corresponding set of candidate context segments $C_T$, we concatenate the input sequence $X$ with a special separator token $\texttt{<|SEP|>}$:
% \begin{equation}
% \label{eq:sft_input_structured}
% X = AudioEncoder(Q_A) \bigoplus_{i=1}^{k} (Tokenizer(c_i) \oplus \texttt{<|SEP|>}).
% \end{equation}
\begin{equation}
\label{eq:sft_input_structured}
\begin{split}
% X = & \, AudioEncoder(Q_A)  \, \bigoplus_{i=1}^{k} (Tokenizer(c_i) \oplus \texttt{<|SEP|>}).
X = E_{\text{audio}}(Q_A) \bigoplus_{i=1}^{k} \left( \text{T}(c_i) \oplus \texttt{<|SEP|>} \right).
\end{split}
\end{equation}
Here, $E_{Audio}$ is the Audio Encoder, $T$ is the Text Tokenizer. %$E_{Audio}$ refers to the Audio Encoder, 

The model is trained to generate an output text sequence $Y$ that consists of the subset of context segments constituting the key evidence. Formally:
\begin{equation}
\label{eq:sft}
Y = SpeechLLM(X).
\end{equation}

The training objective is the standard auto-regressive cross-entropy loss over the target sequence:
\begin{equation}
\mathcal{L}_{SFT} = - \sum_{t=1}^{|Y|} \log P_\theta(y_i | x, y_{<i}; \theta)
\end{equation}
where $y_t$ denotes the $t$-th token in $Y$, and $\theta$ are the model parameters.
\section{Experiment}

\subsection{Environment Setup}
\subsubsection{Datasets and Metrics}
We selected SQuAD v1.1\cite{rajpurkar-etal-2016-squad}, HotpotQA\cite{yang-etal-2018-hotpotqa}, and MuSiQue\cite{trivedi-etal-2022-musique} to cover a broad spectrum of QA capabilities.
We utilized the training sets of these three datasets for LFE training. The specific test set sizes for these datasets are as follows: SQuAD (10,569), HotpotQA (7,404), and MuSiQue (2,417).
Distinct from existing datasets like Spoken SQuAD\cite{li2018spoken} and LibriSQA\cite{zhao2024librisqa} that focus on \textit{Text Query and Spoken Context}, our research targets \textbf{Spoken Query and Text Context}. To address this input modality gap, we synthesized audio queries from standard benchmarks using \textbf{Higgs Audio}\cite{higgsaudio2025}. Evaluation metrics include \textbf{Exact Match (EM)} for answer accuracy and \textbf{Hit Rate}, \textbf{Precision}, \textbf{Recall}, and \textbf{$\text{F}_1$} for evidence grounding.

\subsubsection{Baselines and Base Model}
To evaluate the efficacy of the proposed Attention-guided Evidence Grounding (AEG) method, we established rigorous experimental comparisons. We assessed the generalizability and robustness of AEG by interfacing it with a suite of generative SpeechLLMs, including GPT-4o Audio\cite{gpt4o}, Qwen3-Omni Flash\cite{xu2025qwen3omnitechnicalreport}, Qwen3-Omni-30B-A3B\cite{xu2025qwen3omnitechnicalreport}, and LongCat-Flash-Omni\cite{team2025longcat}. The Qwen2-Audio-7B\cite{qwen2audio} served as the base model for the evidence grounding component. We compared the following experimental configurations:
\begin{itemize}
\item \textbf{Baseline:} The generative model receives the original question speech and the unprocessed context.
\item \textbf{AEG w/o LFE (Ablation):} Omits the LFE training. Evidence grounding is performed using attention scores from the original Qwen2-Audio-7B model.
\item \textbf{AEG with LFE (Full Method):} Fine-tunes the Qwen2-Audio-7B model using the full training sets.
\end{itemize}

\subsubsection{Implementation Details}
\label{sec:implementation}
LFE utilized training splits from SQuAD, HotpotQA, and MuSiQue. Each audio query was paired with $N=10$ candidate passages (ground-truth and negatives) to enforce discrimination. We fine-tuned Qwen2-Audio-7B using AdamW (cosine schedule, LR $1e-5$, warm-up 0.05). Using a batch size of 2 with 8 accumulation steps, we trained for 2 epochs on 8 NVIDIA A100 (80GB) GPUs.

\subsection{Main Results}

We conducted a rigorous evaluation on the Spoken QA dataset. Our experiments benchmarked three distinct methods across three Speech Large Language Models (SpeechLLMs).

\begin{figure*}[t]
\centering
\includegraphics[width=0.90\textwidth]{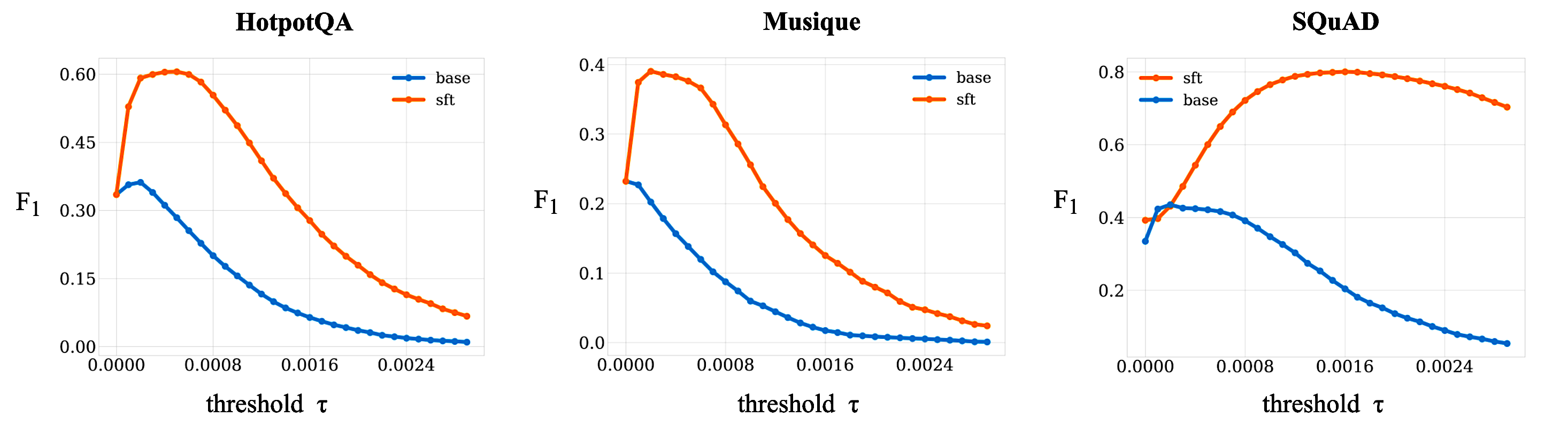}
\caption{Impact of the grounding threshold $\tau$ on evidence selection performance ($\text{F}_1$), comparing the baseline (base) and LFE (sft) models.} \label{fig:ablation_threshold}
\end{figure*}

\begin{table}[h]
\centering
\caption{EM (\%) comparison of methods on context-based Spoken QA tasks across three SpeechLLMs. Best results are bolded.}
\label{tab:model_strategy_comparison}
\resizebox{0.48\textwidth}{!}{
\begin{tabular}{llccc}
\toprule
\textbf{Model} & \textbf{Method} & \textbf{HotpotQA} & \textbf{MuSiQue} & \textbf{SQuAD} \\
\midrule
\multirow{3}{*}{GPT-4o Audio} 
 & baseline & 79.07 & 51.51 & 88.49 \\
 & \textbf{AEG (w/o LFE)} & 78.03 & 53.70 & 86.94 \\
 & \textbf{AEG (with LFE)} & \textbf{79.16} & \textbf{53.99} & \textbf{88.94} \\
\midrule
\multirow{3}{*}{Qwen3-Omni Flash} 
 & baseline & 73.44 & 44.72 & 79.51 \\
 & \textbf{AEG (w/o LFE)} & 73.02 & 44.93 & 80.36 \\
 & \textbf{AEG (with LFE)} & \textbf{74.17} & \textbf{45.72} & \textbf{80.92} \\
\midrule
\multirow{3}{*}{Qwen3-Omni-30B-A3B} 
 & baseline & 75.02 & 45.88 & 88.37 \\
 & \textbf{AEG (w/o LFE)} & 76.69 & 47.62 & 88.96 \\
 & \textbf{AEG (with LFE)} & \textbf{76.95} & \textbf{48.61} & \textbf{89.24} \\
\midrule
\multirow{3}{*}{LongCat-Flash-Omni} 
 & baseline & 74.36 & 49.57 & 84.32 \\
 & \textbf{AEG (w/o LFE)} & 75.34 & 52.38 & 86.61 \\
 & \textbf{AEG (with LFE)} & \textbf{76.09} & \textbf{53.99} & \textbf{87.07} \\
\bottomrule
\end{tabular}
}
\end{table}

As detailed in \Cref{tab:model_strategy_comparison}, our results demonstrate the universality and robustness of the proposed approach. Specifically, \textbf{AEG (w/o LFE)} consistently outperforms standard baselines, while the integration of LFE in \textbf{AEG (with LFE)} yields further significant gains. This efficacy extends across diverse SpeechLLMs, encompassing both closed-source models (e.g., \textbf{GPT-4o Audio}) and open-source models (e.g., \textbf{LongCat-Flash-Omni}). Quantitatively, on the 30B-parameter Qwen3-Omni-30B-A3B, \textbf{AEG (with LFE)} achieves absolute improvements of 1.93\%, 2.73\%, and 0.87\% over the baseline on HotpotQA, MuSiQue, and SQuAD, respectively. Notably, these gains are even more pronounced on the massive 560B-parameter \textbf{LongCat-Flash-Omni}, reaching 1.73\%, 4.42\%, and 2.75\%. These findings underscore the scalability of our approach, validating its ability to deliver performance improvements across model sizes ranging from lightweight architectures to massive SpeechLLMs.

\subsection{Ablation Study}

To thoroughly validate the contributions of our proposed components, we conduct a series of ablation studies.

% Table 3: Using table* for wide content

\subsubsection{Effect of "Learning to Focus on Evidence"}
\label{sec:effect_of_LFE}

We compare our \textbf{AEG with LFE} (full method) against a strong baseline, \textbf{AEG (w/o LFE)}, which omits this training stage. We evaluate both models on the evidence grounding task itself.

% Table 2: Single column using resizebox
\begin{table}[h]
\centering
\caption{Ablation study of the Learning to Focus on Evidence (LFE) module. All metrics are reported in percentages (\%).}
% Comparison between AEG without LFE (w/o LFE) and with LFE (with LFE). 
\label{tab:ablation_sft_all_datasets}
\resizebox{\linewidth}{!}{
\begin{tabular}{llcccc}
\toprule
\textbf{Dataset} & \textbf{Method} & \textbf{Prec.} & \textbf{Rec.} & \textbf{Hit-Rate} & \textbf{$\text{F}_1$} \\
\midrule
SQuAD & AEG (w/o LFE) & 34.21 & 59.68 & 84.93 & 43.49 \\
 & AEG (with LFE) & \textbf{88.07} & \textbf{73.32} & \textbf{91.16} & \textbf{80.02} \\
\midrule
HotpotQA & AEG (w/o LFE) & 26.12 & 58.90 & 81.86 & 36.19 \\
 & AEG (with LFE) & \textbf{56.99} & \textbf{64.57} & \textbf{94.19} & \textbf{60.55} \\
\midrule
MuSiQue & AEG (w/o LFE) & 13.20 & \textbf{96.37} & \textbf{96.77} & 23.22 \\
 & AEG (with LFE) & \textbf{29.21} & 58.83 & 91.06 & \textbf{39.04} \\
\bottomrule
\end{tabular}
}
\end{table}

As shown in \Cref{tab:ablation_sft_all_datasets}, \textbf{AEG (w/o LFE)} struggles to pinpoint evidence, particularly on SQuAD and HotpotQA. In contrast, \textbf{AEG (Ours)} achieves a significant 36.53-point $\text{F}_1$ gain on SQuAD, demonstrating that LFE effectively guides attention focus. Unlike text-based tasks (e.g., Self-Elicit \cite{Liu2025SelfElicitYL}) where explicit training is less critical, the more challenging cross-modal speech domain requires our LFE method for optimal performance.
\Cref{fig:attention_evolution} (a) and (b) provide a qualitative visualization of this process. The heatmaps show that the attention weights on \textbf{key evidence} start diffuse and low, but as training progresses, they become significantly stronger and more consolidated.

\subsubsection{Analysis of Attention Layer Selection}
\label{sec:attention_layer_analysis}

\Cref{fig:attention_evolution} (c) illustrates the optimization of attention layer selection. A strong, positive difference emerges and rapidly concentrates within the middle-to-upper layers, specifically layers 10-28, and remains stable upon convergence. This validates our approach of aggregating attention from this specific range.

\subsubsection{Sensitivity to Grounding Threshold $\tau$}

\Cref{fig:ablation_threshold} illustrates the model performance in terms of \textbf{$\text{F}_1$} score under varying thresholds. Notably, across all three datasets, our \textbf{LFE (sft)} model achieves a significantly higher peak $F_1$ score compared to the \textbf{baseline (base)} model, demonstrating its superior potential in achieving optimal grounding performance.

\subsection{Comparison with Cascade Systems}
\label{sec:cascade_comparison}

Our End-to-End  approach is both faster and more robust than cascade pipelines. We benchmark AEG against cascade systems that pair strong ASR models (Whisper \cite{radford2023robust}) with dense rerankers (BGE-reranker-v2-m3 \cite{chen2024bge} and Qwen3-Reranker \cite{zhang2025qwen3}). Specifically, all models were deployed using vLLM\cite{kwon2023efficient} for efficient inference. Results on SQuAD are reported in \Cref{tab:main_comparison}.

% Table 4: Improved Structure for Cascade Comparison
\begin{table}[h]
\centering
\caption{Performance of Cascade Systems vs. AEG. Metrics are in \% (Latency in ms). Best results are \textbf{bold}.}
\label{tab:main_comparison}
\resizebox{0.48\textwidth}{!}{%
\begin{tabular}{lccccc}
\toprule
\textbf{Configuration} & \textbf{Params} & \textbf{WER} & \textbf{$\text{F}_1$} & \textbf{Hit-Rate} & \textbf{Latency} \\ 
\midrule
\multicolumn{6}{l}{\textit{\textbf{Cascade Systems (ASR + Reranker)}}} \\
\cmidrule(l){1-6}
Whisper-Small + bge-reranker-v2-m3 & 0.8B & 4.71 & 73.85 & 87.53 & 405 \\
Whisper-Medium + bge-reranker-v2-m3 & 1.3B & 3.84 & 74.17 & 87.96 & 503 \\
Whisper-Large-v3 + bge-reranker-v2-m3 & 2.1B & \textbf{3.18} & 74.56 & 88.36 & 599 \\ 
\cmidrule(l){1-6} 
Whisper-Small + Qwen3-Reranker-4B & 4.2B & 4.71 & 73.56 & 88.97 & 420 \\
Whisper-Medium + Qwen3-Reranker-4B & 4.8B & 3.84 & 73.88 & 87.32 & 518 \\
Whisper-Large-v3 + Qwen3-Reranker-4B & 5.6B & \textbf{3.18} & 74.25 & 89.94 & 614 \\ 
\cmidrule(l){1-6} 
Whisper-Small + Qwen3-Reranker-8B & 8.2B & 4.71 & 78.29 & 90.03 & 431 \\
Whisper-Medium + Qwen3-Reranker-8B & 8.8B & 3.84 & 78.91 & 90.90 & 529 \\
Whisper-Large-v3 + Qwen3-Reranker-8B & 9.6B & \textbf{3.18} & 79.14 & 90.11 & 625 \\ 
\midrule
\multicolumn{6}{l}{\textit{\textbf{End-to-End Method}}} \\
\cmidrule(l){1-6}
\textbf{AEG (with LFE)} & 8.2B & N/A & \textbf{80.02} & \textbf{91.16} & \textbf{238} \\ 
\bottomrule
\end{tabular}%
}
\end{table}

\textbf{Robustness against ASR Errors:}

Cascade systems suffer from error propagation: Whisper-Large-v3 still yields a WER of 3.18\%, which degrades downstream reranking. AEG instead ranks directly from audio embeddings in the latent space, avoiding transcription-induced information loss. As a result, AEG reaches an $\text{F}_1$ Score of \textbf{80.02\%} and a Hit-Rate of \textbf{91.16\%}, outperforming even the most expensive cascade setting (Whisper-Large-v3 + Qwen3-Reranker-8B).

\textbf{Efficiency and Latency:}

Cascade pipelines run two large models sequentially (ASR + reranker), leading to high latency (400--600\,ms+). AEG is a lightweight, plug-and-play module: it reuses attention computations from the SpeechLLM prefill phase, so evidence extraction adds minimal computational overhead. As shown in \Cref{tab:main_comparison}, AEG achieves \textbf{238\,ms} average latency, significantly lower than cascade pipelines, making it more suitable for real-time use.

\section{Conclusion}
We proposed \textbf{Attention-guided Evidence Grounding (AEG)} to tackle factual inaccuracy and interpretability issues in Spoken QA. To mitigate the diffuseness of raw attention, we introduced \textbf{Learning to Focus on Evidence (LFE)}, a fine-tuning paradigm that sharpens the model's focus on key evidence. Experiments across SQuAD, HotpotQA, and MuSiQue confirm that our framework consistently yields superior answer accuracy.

% References
% \bibliographystyle{IEEEbib}
\bibliographystyle{IEEEtran}
% \bibliography{icme2026references}

\section{Appendix}
\label{sec:appendix}

\begin{table*}[bp] % 使用 stfloats 后的底部对齐
\centering
\small % 适当缩小字号让版面更精致
\caption{Prompt Template for Spoken QA}
\label{tab:prompt}
\begin{tabularx}{\linewidth}{X} % 使用 tabularx 自动调整宽度
\toprule[1.5pt]
\textbf{System Prompt} \\ \midrule
You are a helpful assistant. Your main job is to understand and refer to the doc in the context to accurately answer questions from users. \\
\\
Reference doc is:\\
\quad doc\_1: doc\_1 content \\
\quad \textcolor{blue}{\textit{$<$EVIDENCE$>$}} \\
\quad doc\_2: doc\_2 content \\
\quad \textcolor{blue}{\textit{$<$/EVIDENCE$>$}} \\
\quad \quad \quad $\vdots$ \\
\quad doc\_n: doc\_n content \\
\\
Please refer to the doc to answer the user's question. You should focus on the doc between the \textcolor{blue}{\textit{$<$EVIDENCE$>$}} and \textcolor{blue}{\textit{$<$/EVIDENCE$>$}} tags, as it contains key information. \\
\\
Directly answer the user's questions, keep the answers as concise as possible, and do not output any irrelevant content. \\
\midrule
\textbf{User Prompt} \\ \midrule
\texttt{<Audio>} [Acoustic Features] \\
\midrule
\textbf{Model Response} \\ \midrule
\texttt{<Answer>} [SpokenQA Answer] \\
\bottomrule[1.5pt]
\end{tabularx}
\end{table*}

\subsection{Prompt Templates}
% \Cref{tab:prompt} shows the prompt template used for the response generation stage of Speech LLM. It instructs the SpeechLLM to answer the acoustic query strictly based on the grounded evidence identified in the previous step, thereby ensuring factual consistency.
\Cref{tab:prompt} presents the prompt template used in the response generation stage of our \textbf{Attention-guided Evidence Grounding (AEG)} framework. After the key evidence grounding step, in which query-relevant context segments are explicitly identified and marked, the prompt instructs the SpeechLLM to generate answers strictly based on the grounded evidence. 
This design reduces the likelihood of hallucinations and ensures that the generated response remains faithful to the identified supporting context.
%This design prevents the model from hallucinating information outside the selected evidence and ensures that the generated response remains faithful to the identified supporting context.

% \begin{table}[htbp]
% \centering
% \caption{Prompt Template for Spoken QA}
% \label{tab:prompt}
% \begin{tabular}{|p{0.9\linewidth}|}
% \hline
% \textbf{System Prompt:} \\
% You are a helpful assistant, your main job is to understand and refer to the doc in the context to accurately answer questions from users;

% Reference doc is:\\
% \\
% doc\_1 : doc\_1 content\\
% \textcolor{bluehighlight}{\textit{$<$EVIDENCE$>$}} \\
% doc\_2 : doc\_2 content \\
% \textcolor{bluehighlight}{\textit{$<$/EVIDENCE$>$}} \\
% \;\;\;\;\;\;\;\;\;\; $\cdots$ \\
% doc\_n : doc\_n content \\
% \\
% Please refer to the doc to answer the user's question. 

% You should focus on the doc between the \textit{$<$EVIDENCE$>$}
% and \textit{$<$/EVIDENCE$>$} tags, as it contains key information.
% \\
% You need to directly answer the user's questions, keep the answers as concise as possible, and do not output any irrelevant content.\\
% \hline
% \textbf{User Prompt:} \\
% $<$Audio$>$ [Acoustic Features] \\
% \hline
% \textbf{Model Response:} \\
% \textbf{Answer:} [SpokenQA Answer] \\
% \hline
% \end{tabular}
% \label{tab:prompt}
% \end{table}

\subsection{Performance Metrics Visualization}

\begin{figure*}[htbp]
\centering
\includegraphics[width=0.9\textwidth]{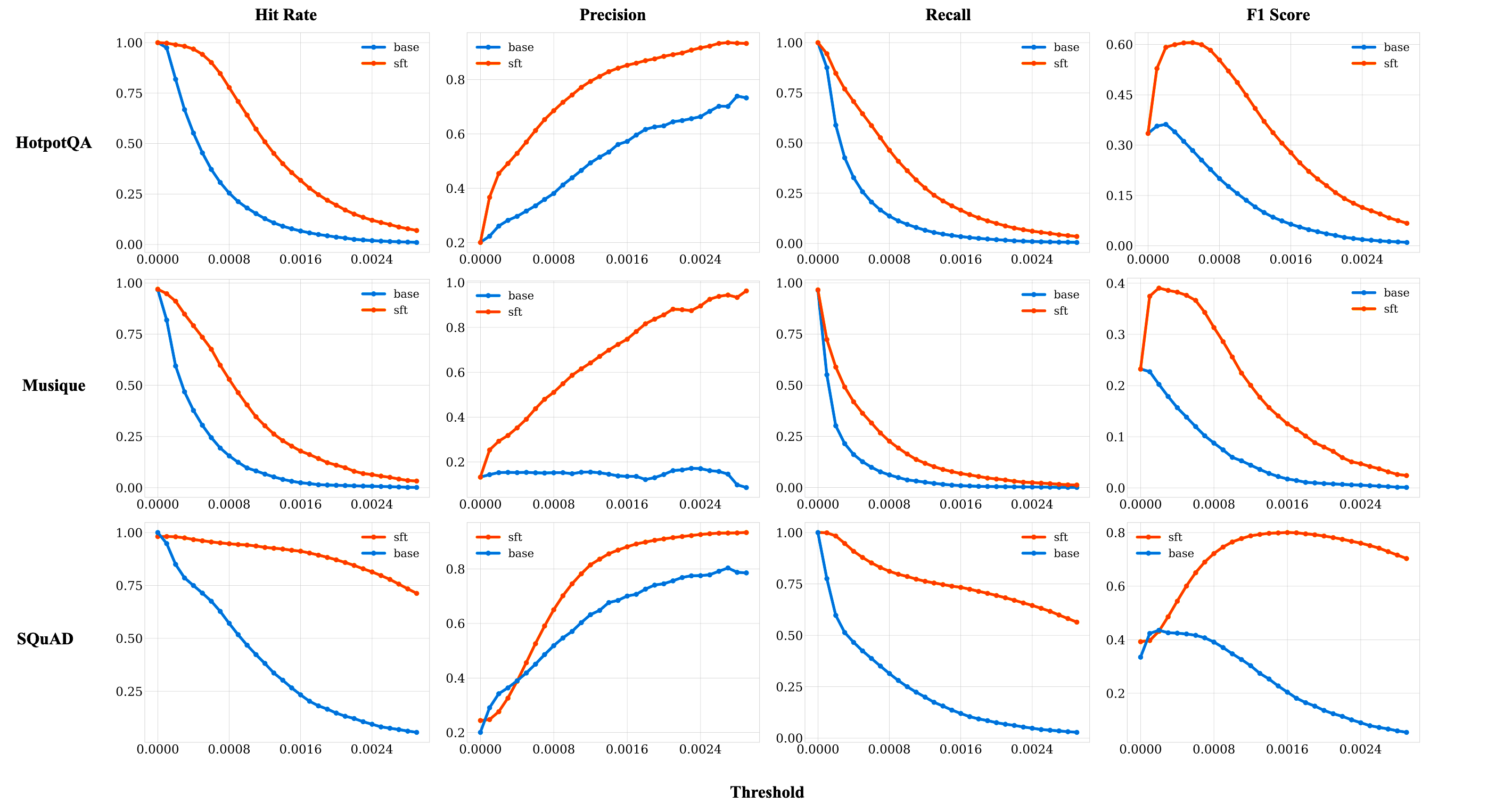}
\caption{Impact of the grounding threshold $\tau$ on evidence selection performance, including Hit Rate, Precision, Recall, and $F_1$ score, comparing the baseline (base) and LFE (sft) models.}
\label{fig:metrics}
\end{figure*}

\Cref{fig:metrics} presents evidence selection performance under varying grounding thresholds $\tau$, evaluated using \textbf{Hit Rate}, \textbf{Precision}, \textbf{Recall}, and \textbf{$F_1$ score}. Across all three datasets (HotpotQA, MuSiQue, and SQuAD), our proposed \textbf{Learning to Focus on Evidence (LFE)} consistently outperforms the \textbf{baseline (base)} model on all metrics, demonstrating its improved ability to balance precision and recall for effective evidence grounding.

%The final evidence selection step relies on a threshold $\tau$ to binarize the attention scores. We analyze the model's sensitivity to this hyper-parameter by varying its value and observing the impact on Precision, Recall, and F1-score.
Evidence selection is performed by applying a threshold $\tau$ to binarize the attention scores. To assess robustness, we vary $\tau$ and analyze its impact on model performance. As $\tau$ increases, Precision rises due to more evidence selection, while Recall and Hit Rate decrease as fewer segments are retained.

% Figure \ref{fig:metrics} illustrates the classic trade-off between Precision and Recall. As $\tau$ increases, Precision rises (the model is more selective) while Recall and Hit Rate decrease (fewer segments are selected, potentially missing some correct ones). 
Importantly, across the entire range of threshold values, our \textbf{LFE (sft)} model (red line) consistently and significantly outperforms the \textbf{baseline (base)} model (blue line) on all four metrics. The F1-score, which balances this trade-off, demonstrates the robustness of our LFE model. It not only achieves a much higher peak F1-score than the baseline but also maintains high performance over a broader range of $\tau$ values. This indicates that our LFE (sft) model is less sensitive to the exact choice of this hyper-parameter, as its learned attention weights provide a more reliable and distinct signal for grounding.

% % References
% \bibliographystyle{IEEEtran}
% \bibliography{icme2026references}
\subsection{Case Study} 
% 1. 挑选一两个有代表性的测试样本，进行案例分析。
% 2. 展示你的模型和基线模型的输出结果，直观地对比优劣。
% 3. 可以分析你的模型为什么在这个案例上表现得更好，例如可以可视化注意力权重等。

\begin{table*}[bp]
\centering
\caption{Examples of how AEG helps LLM, with blue text highlighting the key evidence selected by AEG.}
\label{tab:case_study}
\scalebox{0.95}{
\begin{tabular}{m{11 cm} m{4 cm}}
% \begin{tabularx}{\textwidth}{X m{6cm}} 
\toprule
\multicolumn{1}{c}{\textbf{(Partial) Context Passage}} & \textbf{Question \& Answers} \\
\midrule

% 2. 在 \rotatebox 之前添加 \centering
% \centering\rotatebox{90}{\textbf{True or False}} &
% 单元格 2: Context
doc\_1 Sacro Cuore di Cristo Re is a Roman Catholic church (minor basilica) in Rome, designed between the 1920s and 1930s by Marcello Piacentini. \textcolor{bluehighlight}{$<$EVIDENCE$>$ doc\_2 The post of Governor of Vatican City (Governatore dello Stato della Città del Vaticano in Italian) was held by Marchese Camillo Serafini from the foundation of the state in 1929 until his death in 1952. No successor was appointed, and the post itself was not mentioned in the Fundamental Law of Vatican City State issued by Pope John Paul II on 26 November 2000, which entered into force on 22 February 2001. $<$/EVIDENCE$>$}... ...doc\_19 Bacarra Church is a Roman Catholic church located in the municipality of Bacarra, Ilocos Norte, Philippines under the jurisdiction of the Roman Catholic Diocese of Laoag. The church was founded by the Augustinians, who dedicated it to St. Andrew. doc\_20 Start and end dates vary with location and year... ... & 
% 单元格 3: Q&A
\textbf{Question:} When was the position of Governor ended in the city that houses the head of Catholicism, and the basilica of the saint holding a knife in the Last Supper?

\textbf{True Answer:} 1952.

\underline{\textbf{Baseline}:} 22 February 2001. 

\underline{\textbf{AEG(w/o LFE)}:} 22 Feb-ruary 2001.

\underline{\textbf{AEG(with LFE)}:} 1952  \\
\midrule

doc\_1 The Bellas are left in disgrace after the incident with no hopes of winning the competition until Stacie calls the girls to inform she has given birth to a baby girl and named her Bella. This motivates the Bellas to just perform their hearts out without trying to win... ...\textcolor{bluehighlight}{$<$EVIDENCE$>$ doc\_7 The Battle of New Orleans was a series of engagements fought between December 14, 1814 and January 18, 1815, constituting the last major battle of the War of 1812. American combatants, commanded by Major General Andrew Jackson, prevented a much larger British force, commanded by Admiral Alexander Cochrane and General Edward Pakenham, from seizing New Orleans and the vast territory the United States had acquired with the Louisiana Purchase.$<$/EVIDENCE$>$} doc\_8 The Battle of Alto de los Godos was a battle that took place on 25 ... ... doc\_20 The Battle of Alexandria or Battle of Canope ... ... &
% 单元格 3: Q&A
\textbf{Question:} Who was the British general in the birthplace of the performer of Here Come the Girls?

\textbf{True Answer:} Edward Pakenham.

\underline{\textbf{Baseline}:} William Howe. 

\underline{\textbf{AEG(w/o LFE)}:} Based on the given information, the question cannot be answered.

\underline{\textbf{AEG(with LFE)}:} Edward Pakenham. \\
\bottomrule

\end{tabular}

}
\end{table*}

% For qualitative illustration, in Table \ref{tab:case_study} we show two examples of complex questions. AEG demonstrates its ability to identify the crucial supporting facts from a larger context, guiding the model to the correct answer. For instance, in the first example asking for the end date of the Governor of Vatican City, AEG highlights the critical evidence in doc\_2 stating the post was held ``until his death in 1952''. This prevents the model from being misled by other dates in the passage (like ‘22 February 2001’), which confused the model in baseline. In the second multi-hop example, which asks for a British general, AEG pinpoints the overlooked doc\_7 as the key evidence. This snippet explicitly names ‘General Edward Pakenham’, guiding the model arrive at the correct answer, whereas the base model failed and the AEG variant without LFE abstained.
To qualitatively analyze model behavior, Table~\ref{tab:case_study} presents two representative examples involving complex, multi-fact queries. These instances underscore AEG's capacity to discern pivotal supporting facts within a broad context, thereby guiding the inference process toward the correct answer.

In the first example, which requires determining the tenure of the Governor of Vatican City, AEG accurately isolates the key evidence in \texttt{doc\_2} (``until his death in 1952''). 
This precise extraction mitigates irrelevant temporal information in the context, such as `22 February 2001', which misleads the baseline model and results in an incorrect prediction.

In the second scenario, which queries the identity of a British general, AEG demonstrates its strength by pinpointing \texttt{doc\_7}, a document overlooked by the baseline model. 
By highlighting the doc that explicitly mentions `General Edward Pakenham', AEG derives the correct answer. 
In contrast, the baseline model fails to highlight this evidence, while the AEG variant without LFE abstains from answering. 
This case highlights the importance of of the full architecture in calibrating attention distributions, particularly for low-resource or less salient queries where relevant evidence is sparse or weakly signaled.

\end{document}